\documentclass[english]{IEEEtran}

\usepackage{xparse,xcolor}

\ExplSyntaxOn

\NewDocumentCommand{\setupbibcolors}{m}
 {
  \cs_set_protected:Npn \bibitem ##1
   {
    \color{ \str_case:nnF { ##1 } { #1 } { black } }
    \heba_bibitem:n { ##1 }
   }
 }

\cs_set_eq:NN \heba_bibitem:n \bibitem

\ExplSyntaxOff

\setupbibcolors{
  {Baza3}{white}
  {CC3}{white}
  {baza2019b}{white}
  {parksmarnet}{white}
   {parkccnc}{white}
  {BC2}{white}
  {baza2019detecting}{white}
  {firmware2}{white}
   {baza2019sharing}{white}
    {baza2019blockchain}{white}
    {omar1}{white}
    {omar2}{white}
    {shafee2019mimic}{white}
}

\usepackage[T1]{fontenc}
\usepackage[latin9]{inputenc}
\PassOptionsToPackage{draft}{hyperref}
\usepackage[bookmarks=false]{hyperref}
\PassOptionsToPackage{bookmarks=false}{hyperref}
\usepackage[output-decimal-marker={.}]{siunitx}

\usepackage{caption}

\PassOptionsToPackage{bookmarks=false}{hyperref}
\usepackage{array}

\newcolumntype{C}{>{$}c<{$}}
\usepackage{booktabs}

\newsavebox{\foobox}

\usepackage{bm}

\usepackage{makecell}
\usepackage{tikz}
\usetikzlibrary{shapes.arrows}
\usepackage{color}
\usepackage{xcolor}
\usepackage{array}
\usepackage{booktabs}
\usepackage{textcomp}
\usepackage{multirow}
\usepackage{mathtools}

\usepackage{url}
\usepackage{pifont}
\usepackage{amsfonts}

 \usepackage{indentfirst}
\usepackage{amsthm}
\usepackage{amssymb}
\usepackage[export]{adjustbox}
\usepackage{verbatim}
\usepackage{graphicx}
\usepackage{mathrsfs} 
\DeclareMathAlphabet{\mathpzc}{OT1}{pzc}{m}{it}
\usepackage{booktabs}

\usepackage{color}

\usepackage[ruled,linesnumbered]{algorithm2e}
\definecolor{seagreen}{rgb}{0.18, 0.55, 0.24}
\SetAlFnt{\small\color{black}\tt}
\LinesNumbered

\def\BibTeX{{\rm B\kern-.05em{\sc i\kern-.025em b}\kern-.08em T\kern-.1667em\lower.7ex\hbox{E}\kern-.125emX}}

\newcounter{defcounter}
\usepackage[shortlabels]{enumitem}
\usepackage{cite}
\usepackage[ruled]{algorithm2e}
\mathchardef\period=\mathcode`.
\DeclareMathSymbol{.}{\mathord}{letters}{"3B}
\usepackage{tikz}
\tikzstyle{io} = [fill=black,inner sep=2pt,circle]

\graphicspath{ {images/} }

\usetikzlibrary{arrows,positioning}
\usetikzlibrary{shapes.geometric,positioning,shapes,arrows,calc, decorations.pathreplacing, arrows.meta}

\everymath{\displaystyle}
\usetikzlibrary{backgrounds}

\usepackage{pifont}

\makeatletter
\def\endthebibliography{%
	\def\@noitemerr{\@latex@warning{Empty `thebibliography' environment}}%
	\endlist
}

\makeatletter
\newcommand*\bigcdot{\mathpalette\bigcdot@{.5}}
\newcommand*\bigcdot@[2]{\mathbin{\vcenter{\hbox{\scalebox{#2}{$\m@th#1\bullet$}}}}}
\makeatother

\makeatletter

\theoremstyle{plain}

\usepackage{cite} 
\usetikzlibrary{shapes,arrows}
\tikzstyle{line}=[draw] 
\usepackage{algorithmic}
\usepackage{graphicx}
\usepackage{enumitem}

\usepackage{url}
\usepackage{soul}
\usepackage{xcolor}

\makeatother
\usepackage{babel}
\providecommand{\theoremname}{Theorem}

\begin{document}

\title {Practical Fast Gradient Sign Attack against
Mammographic Image Classifier
}
\author{\IEEEauthorblockN{Ibrahim Yilmaz}\\
\IEEEauthorblockA{\textit{Department of Computer Science} \\
\textit{Tennessee Technological University}\\
Cookeville, TN, United States of America \\
 iyilmaz42@students.tntech.edu}
}
\maketitle
 \begin{abstract}
Artificial intelligence (AI) has been a topic of major research for many years. Especially, with the emergence of deep neural network (DNN), these studies have been tremendously successful. Today machines are capable of making faster, more accurate decision than human. Thanks to the great development of machine learning (ML) techniques, ML have been used many different fields such as education, medicine, malware detection, autonomous car etc. In spite of having this degree of interest and much successful research, ML models are still vulnerable to adversarial attacks. Attackers can manipulate clean data in order to fool the ML classifiers to achieve their desire target. For instance; a benign sample can be modified as a malicious sample or a malicious one can be altered as benign while this modification can not be recognized by human observer. This can lead to many financial losses, or serious injuries, even deaths. The motivation behind this paper is that we emphasize this issue and want to raise awareness. Therefore, the security gap of mammographic image classifier against adversarial attack is demonstrated. We use mamographic images to train our model then
evaluate our model performance in terms of accuracy. Later on, we poison original dataset and generate adversarial samples that missclassified by the model. We then using structural similarity index (SSIM) analyze similarity between clean images and adversarial images. Finally, we show how successful we are to misuse by using different poisoning factors.

	\end{abstract}
	\begin{IEEEkeywords}
DNN, adversarial ML, security and Security.

	\end{IEEEkeywords}
\section{Introduction}


Cancer is caused by uncontrolled reproduction of some cells in the body, with breast cancer being one of the most prevalent cancer types that occurs among women. Cells reproduce in order for the body to regenerate and heal, but mutations can cause cells to overproduce and create tumors. 

The tumors can be benign or malignant. Benign tumors are considered harmless and appear normal and do not invade the other tissues nearby. On the other hand, malignant tumors are the cancerous and spread other parts of the body.

Breast cancer usually starts to occur in the milk ducts most of the time and commonly in the lobules of the breast \cite{Breast}. Though it is unlikely males also get breast cancer. The reason of the cancer might vary, and usually there is more than one factor including old age, excessive alcohol and tobacco consumption or simply genetic predisposition.

For treatment, early identification is key, otherwise the tumor can lead to loss of the breast for the person. One of the best methods to recognize the the breast cancer is to use mammography. However, poor quality can result in a misdiagnosis. In order to decrease the occurrence of misdiagnosis, ML models have been in the usage in the field of computer science to identify cancerous cells on mammographic images and get more accurate results.

Along with developing ML, ML has become significant separating images of cancerous cell from images of normal cells. Recently, there have been much successful research to diagnose and classify breast cancer using DNNs. Due to the high performance and effective computation, DNNs are greatly preferred regarding classification of mammographic images; according to a research, convolutional neural network (CNN),is a type of DNN, effectively classifies malicious and benign samples on mammogram images \cite {sahiner1996classification,yilmaz2019expansion}. Although CNN is widely used for detection of cancereous cell, it relies on data-driven approaches and the performance of the classifier based on data. Therefore, the user should think carefully and consider these simple questions.

"Is the mammographic image classifier vulnerable to adversarial data manipulation?

Is it easy to find adversarial way to fool mammographic image classifier?"

Unfortunately, the answer of both questions are yes. In that case, how can the robustness of the breast cancer classifier model be trusted if data can be manipulated easily? In this paper, the issue is addressed by showing different case studies and analyze a potential solution regarding this problem.

\subsection{Contributions}

Recent research has demonstrated that CNNs perform poorly on altered images generated by injecting subtle perturbation to original image. These kind of crafted images may change decision of CNN on mammographic image. It leads to a misclassification of notably automated breast cancer detection over given adversarial mammographic image. This situation may cause a misdiagnosis of cancer to healthy person and vice versa. It may result in breast loss or financial and psychological consequences. This paper contains the following
constructive contributions to highlight this security weakness and heighten awareness of physicians when they give a diagnosis of cancer:

\begin{itemize}
\item We highlight and assess the security vulnerability of CNN widely used in the diagnosis of breast cancer.
\item We implement an efficient adversarial attack on mammographic images in order to fool the mammographic image classifier. \textit{To the best of our knowledge}, adversarial attack applied on mamograpgic image classifier is the first time.  

\item  We analyze the comparison between original  and crafted mamograpgic images using SSIM and analyze the how original image change based on different perturbation coefficient.
\end{itemize}

The rest of this paper is organized as follows: the necessary background of CNN was reviewed in Section II. We present the proposed attack strategy on mammographic image in Section III. We implement of attack on CNN models in
Section IV. The evaluation results of our adversarial attack are presented in Section V. We look into the literature review in the context of our work in Section VI. Ultimately, we finalize the paper in section VII.

\section{Background}

 In this section, we briefly go through learning process of DNNs along with a deep learning approach using CNN on the mammographic image.

\subsection{Learning Process}
\par DNNs like humans, have learning process with experience. Available data is the experience of DNNs. DNNs try to capture distribution of training data which helps to determine boundary decisions between the normal and the cancerous images. Built model then  predicts whether the unlabelled image is cancerous or normal.
\par DNN is supervised learning process meaning that a paired training data X and Y is given. Here X represents the mammographic image and Y represents the ground-truth label as cancer or normal. The aim of the model is to find a function f(X) that fits the map perfectly from X to Y. Model takes each mammographic image and their corresponding label data and learns the behaviour of model by minimizing error.

\subsection{Deep Learning Approach Using CNN on Mammographic Image}

\par
\begin{figure}[!ht]
\centering
\includegraphics[width=8cm]{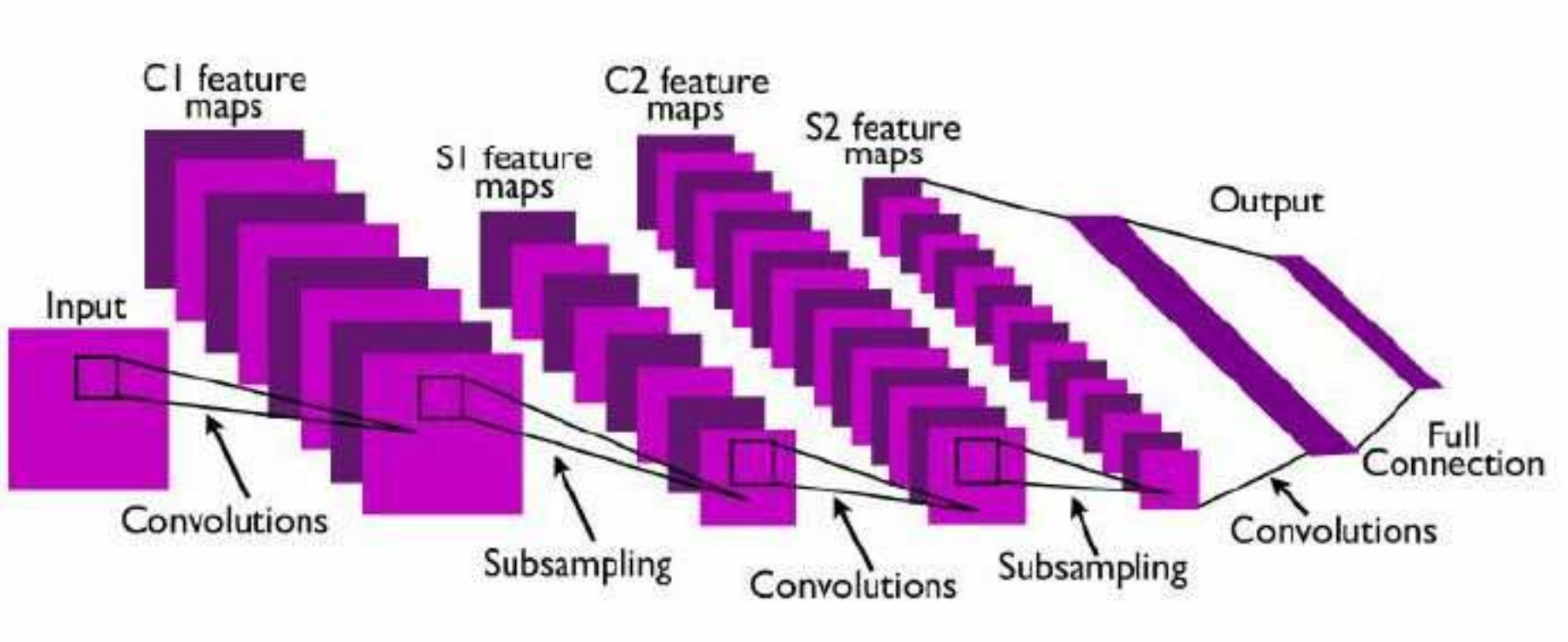}
\caption{ CNN Architecture with Two Phase.}
\label{CNN}
\end{figure}

 CNN is a special type of deep neural network which is widely used for image classification. As DNN, CNN comprises of input, output and hidden layers.  However, hidden layer is comprised of different parts knowns as convolution, pooling and fully connected, different from the  DNN. An example of CNN architecture with two phase is epicted in Figure \ref{CNN}. It is taken in \cite {lewis2013reinforcement}.

Convolutional layer and pooling layer is used for feature selection of the mamographic image, and these feature are connected with each other in the fully connected layer. 

\textbf{\textit{Convolutional layer:}} In this layer, filter matrix which is special matrix and its size depend on architecture model is applied over mammographic image. Convolution operation is a linear operation, so to make it non-linear, activation function is applied following convolution operation. This is because mostly complex problems are nonlinear and can be solved by nonlinear functions.

\textbf{\textit{Pooling Layer:}} Pooling layer is a subsampling process. It is mostly used following convolutional process and there is no learning part in this layer. There are different types of pooling layers such as max pooling, mean pooling, sum pooling etc. Most often, max pooling is used because of its high performance. In the maximum pooling, the maximum pooling function takes group of pixel values as input and chooses the biggest one of this group as output. 

\textbf{\textit{Fully connected layer:}} CNN is completed with this layer. In this layer, all nodes are connected and are used with softmax or sigmoid function in order to predict probabilities of each output through learning process.

\subsection{Structural Similarity Index (SSIM)}

Structural similarity Index is used a lot to compare different images in terms of similarity. Prior to the SSIM, mean square error was preferred by researchers, yet because of its limited functionality, a new image quality assessment was needed. Wang et al. in \cite {wang2002universal} brought a new solution to evaluate image quality assessment in terms of luminance, contrast, and structure.

Luminance is a measure that indicates how much light passes through, along with what is reflected or spread from a certain region at a certain time \cite{Luminance}. However, the simple definition of contrast is differences between lowest and highest pixel density in a image \cite{Contrast}. In addition, structure is another metric used to distinguish images from each other. It is used to tell objects apart. If a image is considered, then it assesses in terms of edges, corners, sub-regions etc of the image. 

Multiplicative function of these three features give a structural similarity index value. If the SSIM value is high, it means that these two images are very similar. If it is low when two images are compared, it is interpreted as different images.

\section{Attack Model}
In this section, we introduce our attack model based on CNN which is inspired in \cite{goodfellow2014explaining} in order to generate adversarial mammographic image samples that fool the mamograpgic image classifier.

ML models are data driven approach and aim to minimize cost function using the gradient descent algorithm. The idea behind this attack is to find a adversarial direction and all pxel values change based on that direction in an attempt to lead to a misclassification. Adding imperceptibly small computed noise on the same direction of cost function is able to maximize cost function instead of minimize it. To put it simply, attackers use gradient cost function with respect to input data and add small amount of perturbation to maximize the cost. By doing this way, attackers craft adversarial samples which are assigned to the incorrect labels. We can explain the mathematical definition of our algorithm below.

Let M be our mammograpgic image classifier model. x is a real mammographic image and y is the corresponding ground-truth output label (cancer or normal) of given the image. C( M,x,y) is the cost function used to train model, $\epsilon$ is the penetration coefficient, \textit{$\hat{x}$} is the adversarial mammographic image is crafted by the adversarial and $\hat{y}$ is the prediction of the model given a adversarial image (\textit{$\hat{x}$}) as an input. 

\begin{equation}
 \hspace{0.3cm} objective \hspace{0.6cm}  max \hspace{0.1cm} c(M,\hat{x},y)
\end{equation}

\begin{equation}
subject \hspace{0.1cm} to \hspace{0.5cm} \hat{x} = x + \delta x 
\end{equation}

\begin{equation}
\left|x\right|_{d} \hspace{0.5cm} \leq \hspace{0.5cm} \epsilon \ast \left|x\right|
\end{equation}

in (3) represents dimensionality of mammographic image data is allowed to be altered. We set d is infnity as proposed in \cite {goodfellow2014explaining}. It means that it is allowed to add perturbation on any dimension of the original mammograpgic images.

In addition, $\delta x$ in (2) denotes perturbation added to the original mamographic images. The magnitude of this perturbation should be equal or less than $\epsilon$ as constrained in (3) where $\epsilon$ controls perturbation amount. The aim is to maximize error in (1).

$\delta x$  is computed as follows: 

\begin{equation}
\hspace{0.3cm} \delta x = \epsilon \ast \hspace{0.1cm} sign(\nabla x \hspace{0.1cm}c(M,x,y))
\end{equation}

where \textit{sign} ($\nabla$x c(M,x,y) )) is the direction of minimizing cost function of the model and $\epsilon$ controls the penetration magnitude in (3).

Therefore, cost function of adversarial mammographic images is  $c(M,\hat{x},y)$ and y $\ne$ $\hat{y}$ must be true if this attack is succeed. 

\begin{figure}[!ht]
\centering
\includegraphics[width=8cm]{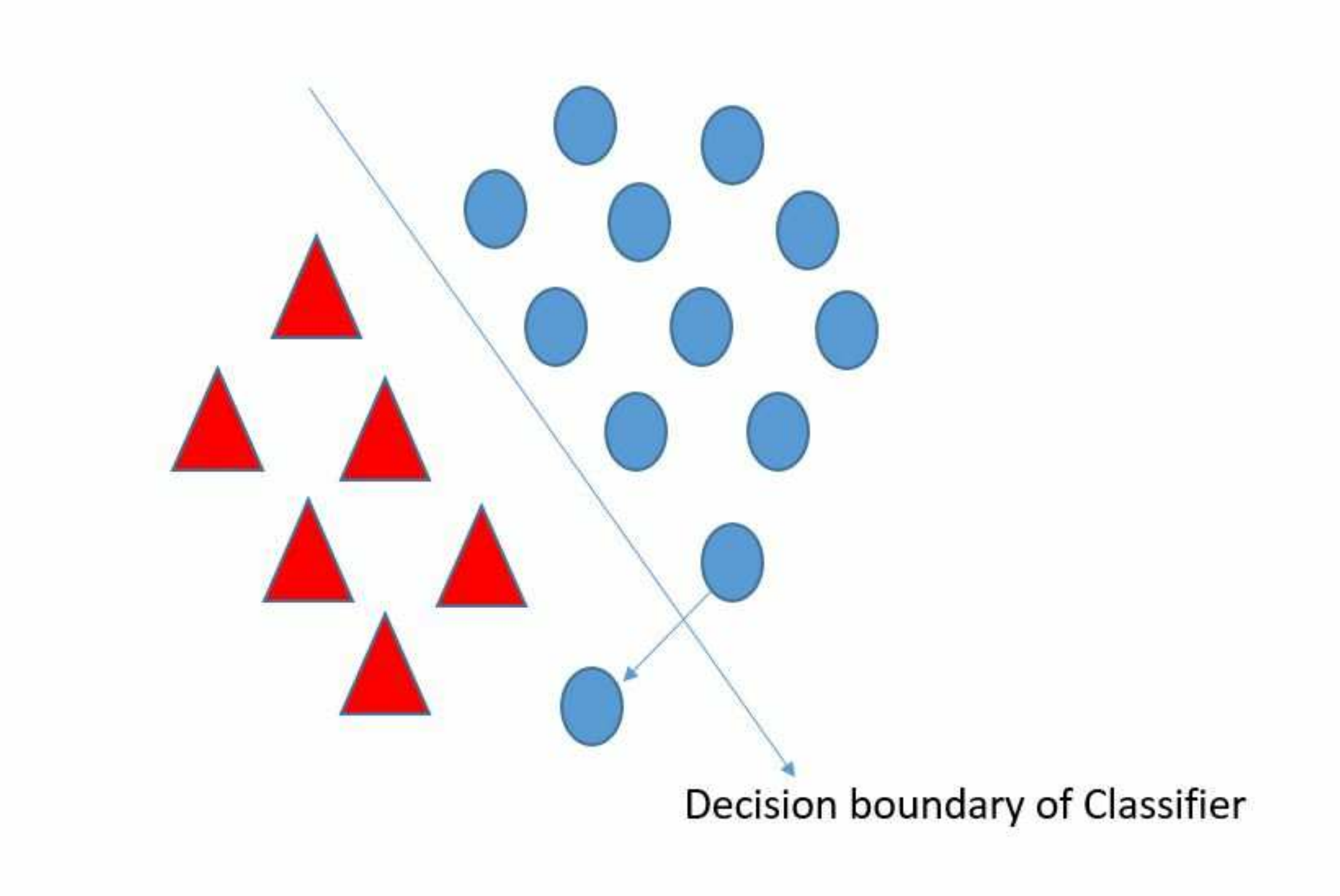}
\caption{ Adversarial Attack Visualization.}
\label{adversary}
\end{figure}

To visualize, we draw our attack scenario in Figure \ref{adversary} with triangulars representing cancer cells and circles representing normal cells. A normal cell which is close to the boundary decision is predicted truly by the CNN model. After adding subtle perturbation on test phase, the same model predicts the cancer cell incorrectly. Because of the limited feature precision, small changes of pixel value are not able to be realized by human eyes. However, these changes may cause the activation function to grow exponentially and fool the breast cancer classifier.

Range of the $\epsilon$ value is between 0 and 1. Rising $\epsilon$ value increases the probability of \textit{$\hat{x}$} being misclassified by the model M. On the other hand, increasing $\epsilon$ value leads to adversarial samples to be detected by human eyes. There is a trade-off between increasing the probablity an attack will be successful and being recognizable by human. In Evaluation Section,  we also evaluate the how much the original images are modified through adversarial attacks using a structural similarity index.

\section{ Material and Method}

\subsection{Data Processing}

The University of South Florid provides mammographic image dataset in order for researchers to develop new techniques for early diagnosis of breast cancer \cite{dataset}. The database includes roughly 2500 samples. However, we could use 100 od them for our work because of limitations. Each sample has ground-truth label as normal or cancer.   

Collection of data from dataset is challenging since the dataset has different files, each file includes different studies, and the image format is 'GIF'. After data has been collected from different files, each image has been converted 'JPG' format manually. Furthermore, each image size is different. Prior to images have been used as input of our model, with the dimensions of the images being set up 256 X 256. 

\subsection{Implementation of Model}

 We encode our implementation in Python using Tensorflow \cite{abadi2016tensorflow} in ourder build our model based on a CNN. Number of convolution layers, pooling layers, fully connected layers and iterations mainly depends on the CNN's architecture and the dataset itself to find optimum parameters. Therefore, we try different architectures until we reach an optimum solution. We train our model with optimum accuracy performance including 2 different convolution layers, each with a 5 X 5 kernel size, two different max pool layers and two different fully connected layers, with 180 X 50 and 50 X 2 sizes respectively.

Afterwards, we apply FGSM attacks to our trained model with different $\epsilon$ values. We produce adversarial samples for each $\epsilon$ value so as to deceive victim model and we monitor how its accuracy changes based on different $\epsilon$  values. We set $\epsilon$  value to 0 as well because it represents the model performance on the original test set. In addition, we monitor how original image changes with different $\epsilon$  values using SSIM technique and try to figure out which epsilon value can be more successful without being recognizable by people.

\section{ Evaluation}

We use 100 mammographic images taken from 'Digital Database for Screening Mammography' (DDSM) database released by the University of South Florida. 70 of the images are obtained from a healthy person and rest of them are taken from cancerous patient. We divide our database in two parts. 90 images are used to train our model and the 10 remaining images are used to test accuracy of our model.

\begin{figure*}[htbp]
  \begin{minipage}[b]{0.48\textwidth}
    \includegraphics[width=\textwidth]{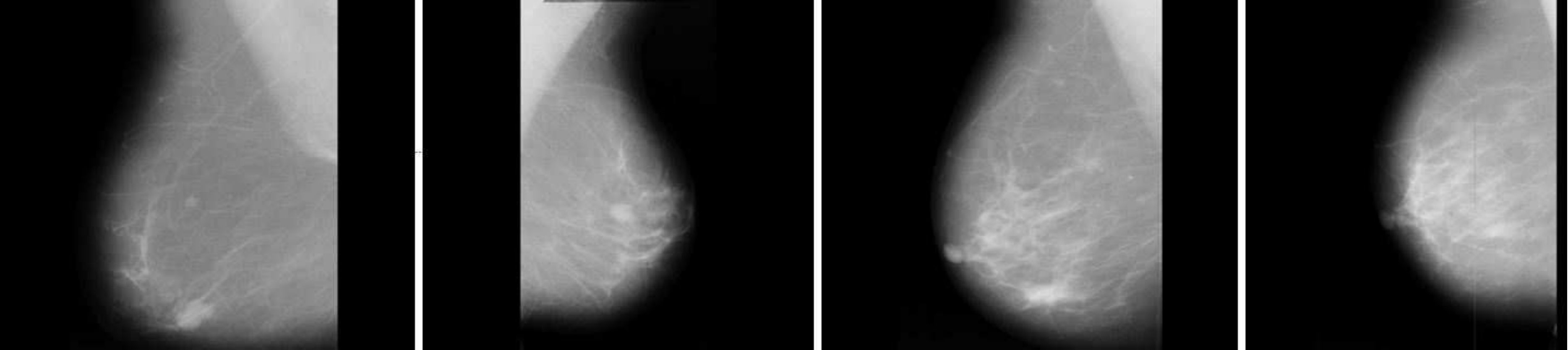}
    \caption{Cancer Images.}
    \label{cancer}
  \end{minipage}\hspace{.3cm}
  \begin{minipage}[b]{.48\textwidth}
    \includegraphics[width=\textwidth]{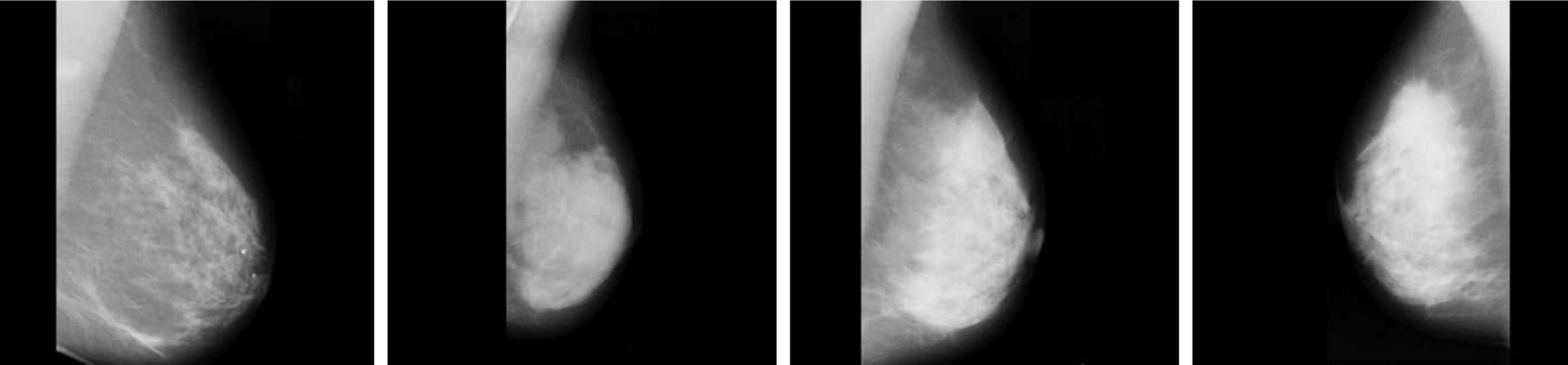}
    \caption{Normal Images.}
    \label{normal}
  \end{minipage}
\end{figure*}

There are some samples of cancer images in Figure \ref{cancer} and normal images in Figure \ref{normal}. We evaluate our attack success by observing decreasing accuracy of the CNN model. Adversarial samples which we generate mislead the model into making incorrect decision. 

\begin{figure*}[htbp]
  \begin{minipage}[b]{0.48\textwidth}
    \includegraphics[width=\textwidth]{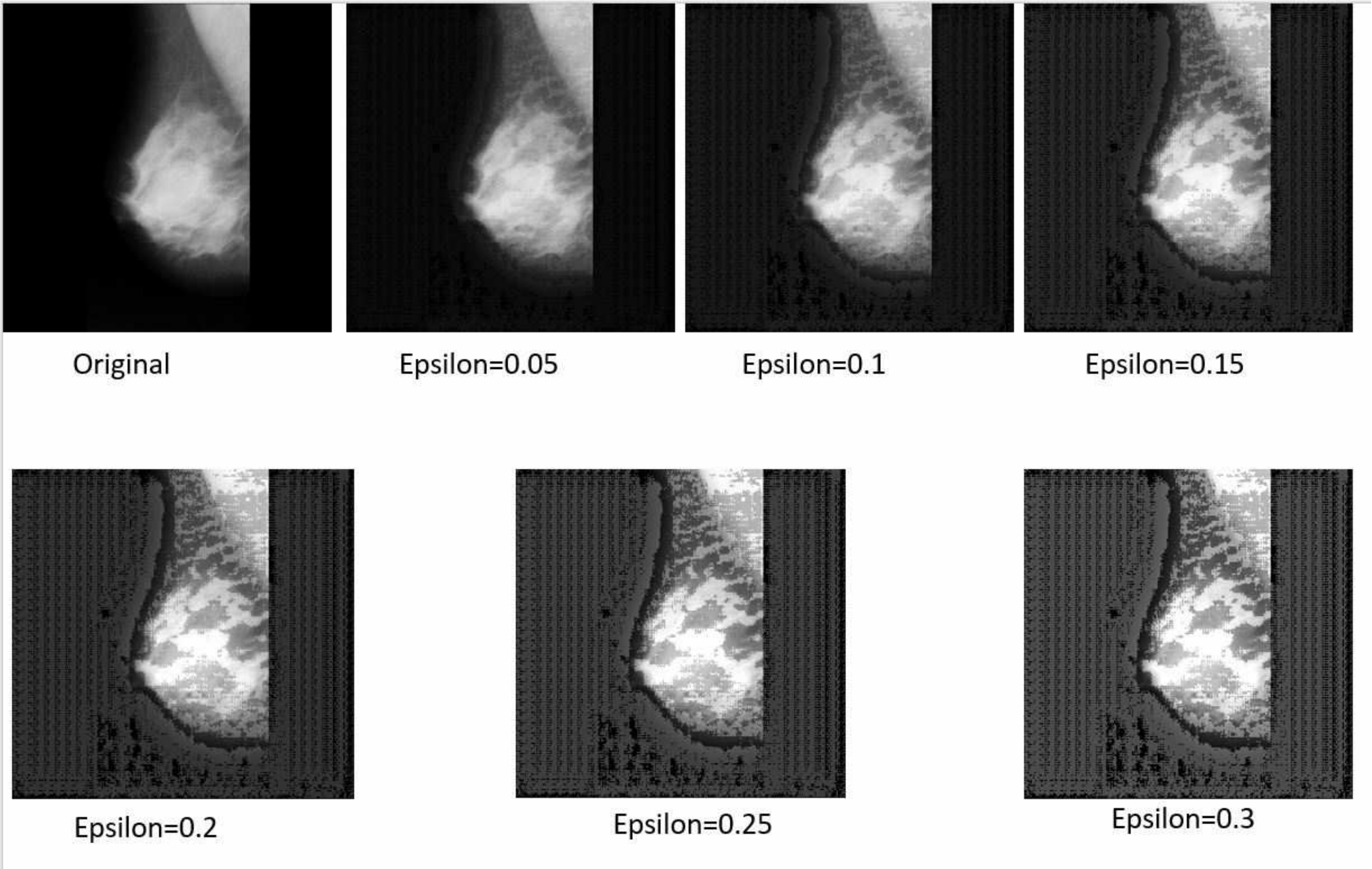}
    \caption{Adversarial Samples for Cancer Images with High Epsilon Values.}
    \label{cancer_adv}
  \end{minipage}\hspace{.3cm}
  \begin{minipage}[b]{.48\textwidth}
    \includegraphics[width=\textwidth]{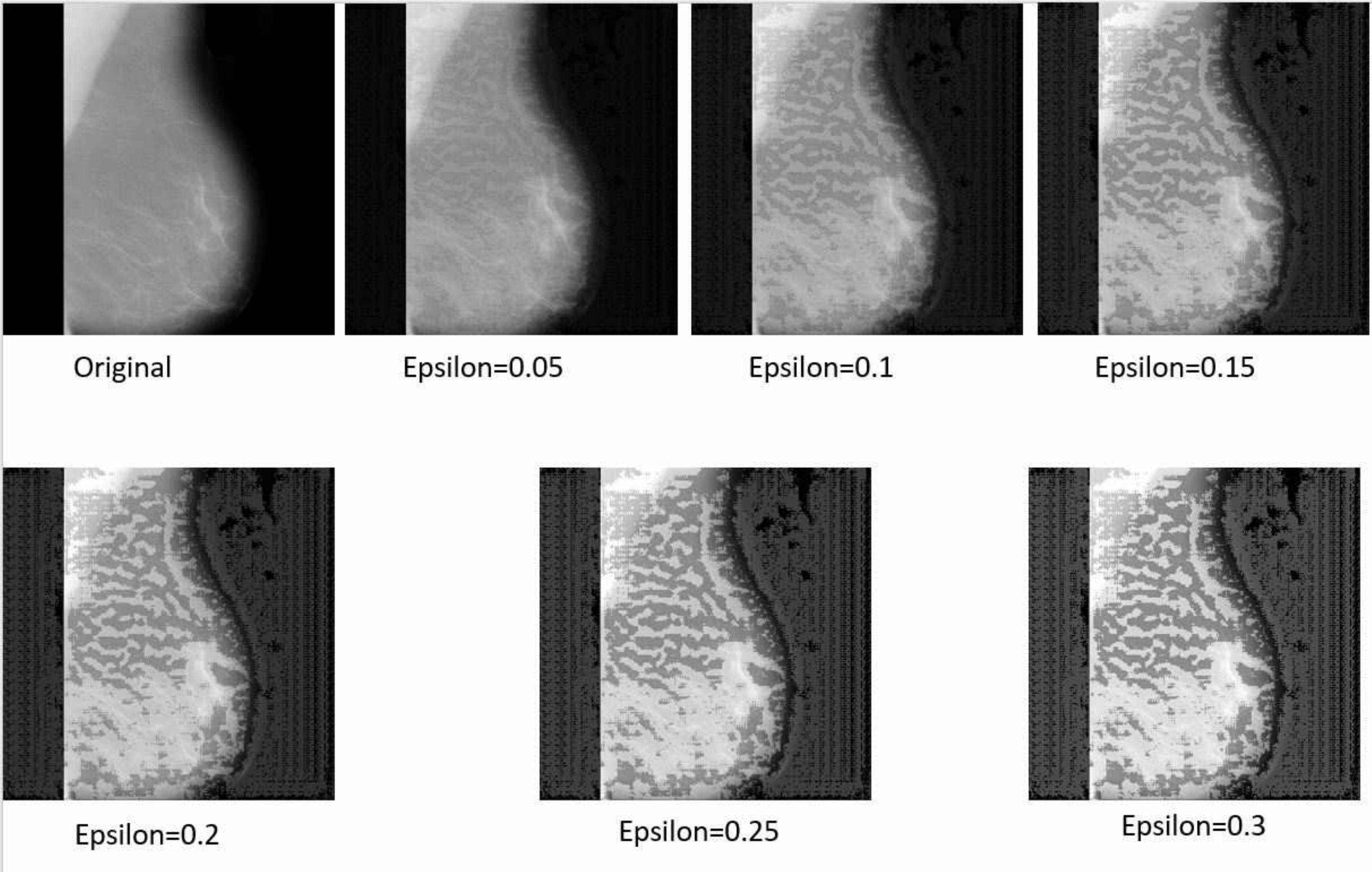}
    \caption{Adversarial Samples for Normal Images with High Epsilon Values.}
    \label{normal_adv}
  \end{minipage}
\end{figure*}

Using FGSM attack algorithm, we craft attack samples for both cancer and normal images. Some are depicted in Figure \ref{cancer_adv} and in Figure \ref{normal_adv} based on different epsilon values.

We sometimes set epsilon values high intentionally to demonstrate maximum damage to victim model and to show how the penetrated image is easily detected by observer with high epsilon values. We modify epsilon values with small values and monitor how much adversarial samples impair performance of our model without being recognizable. Adversarial samples are produced with small epsilon values In Figure 7 and in Figure 8.

\begin{figure*}[htbp]
  \begin{minipage}[b]{0.48\textwidth}
    \includegraphics[width=\textwidth]{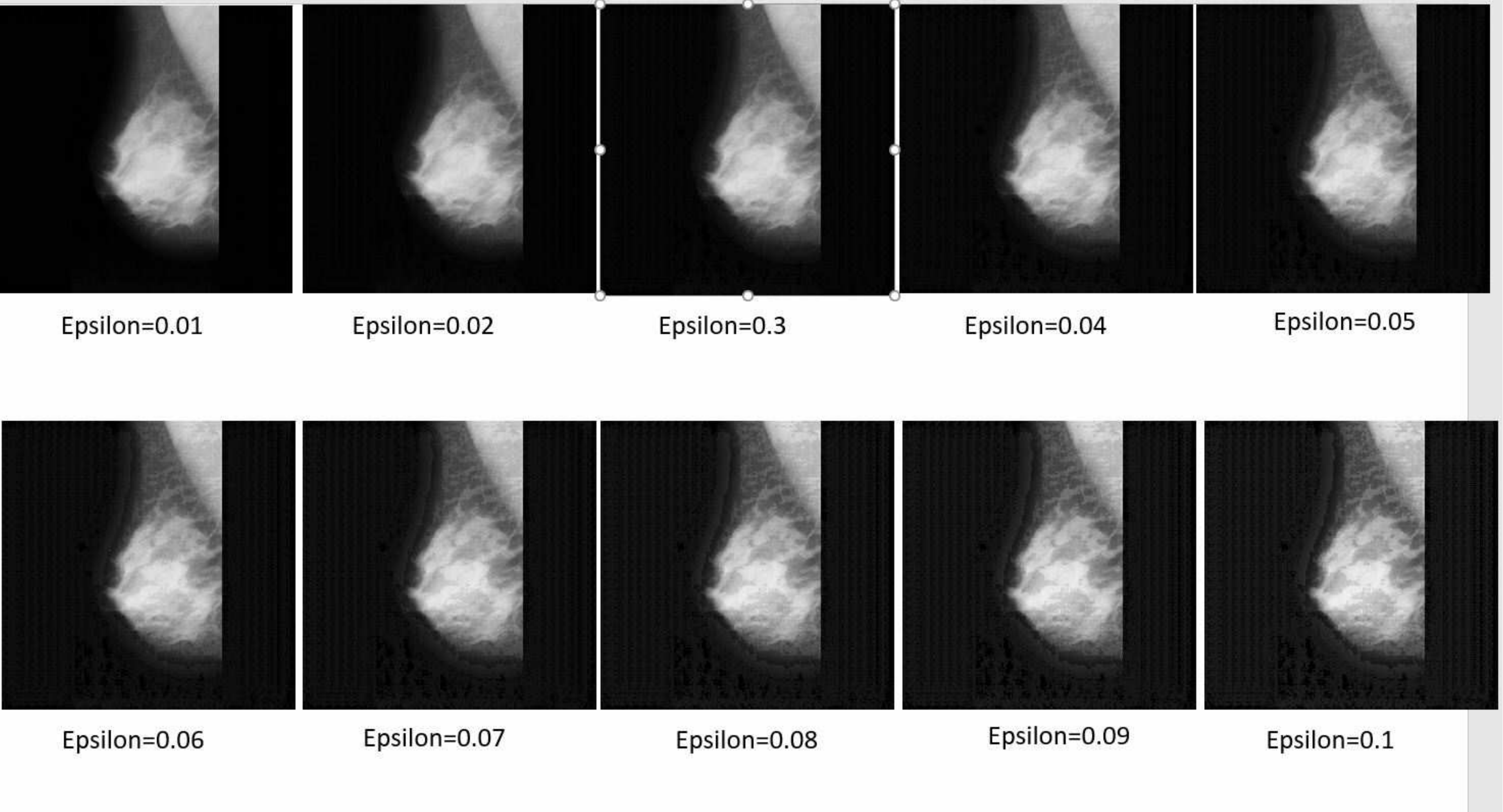}
    \caption{Adversarial Samples for Cancer Images with Small Epsilon Values.}
    \label{cancer_high}
  \end{minipage}\hspace{.3cm}
  \begin{minipage}[b]{.48\textwidth}
    \includegraphics[width=\textwidth]{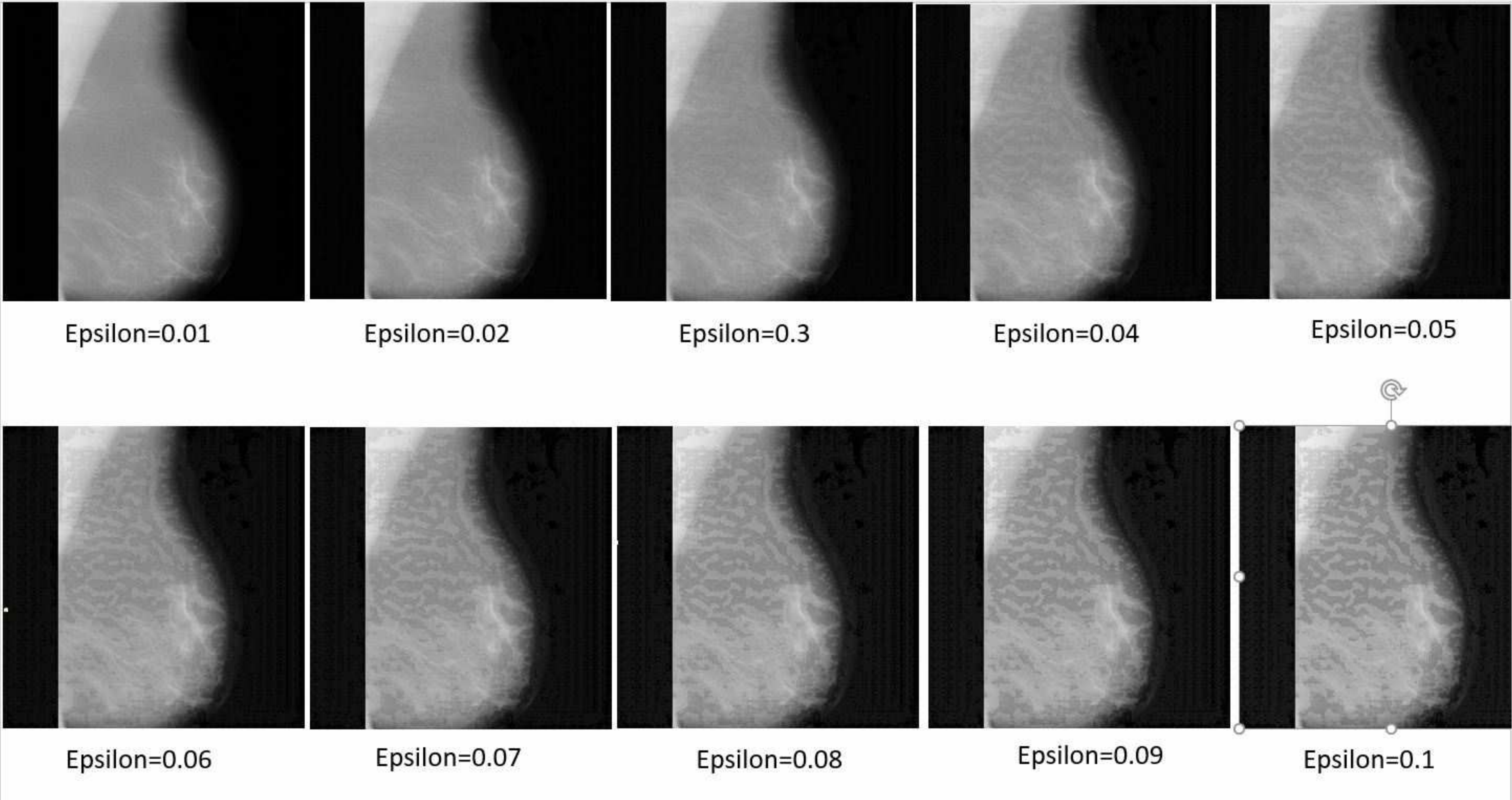}
    \caption{Adversarial Samples for Normal Images with Small Epsilon Values.}
    \label{normal_high}
  \end{minipage}
\end{figure*}

\begin{figure*}[htbp]
  \begin{minipage}[b]{0.48\textwidth}
    \includegraphics[width=\textwidth]{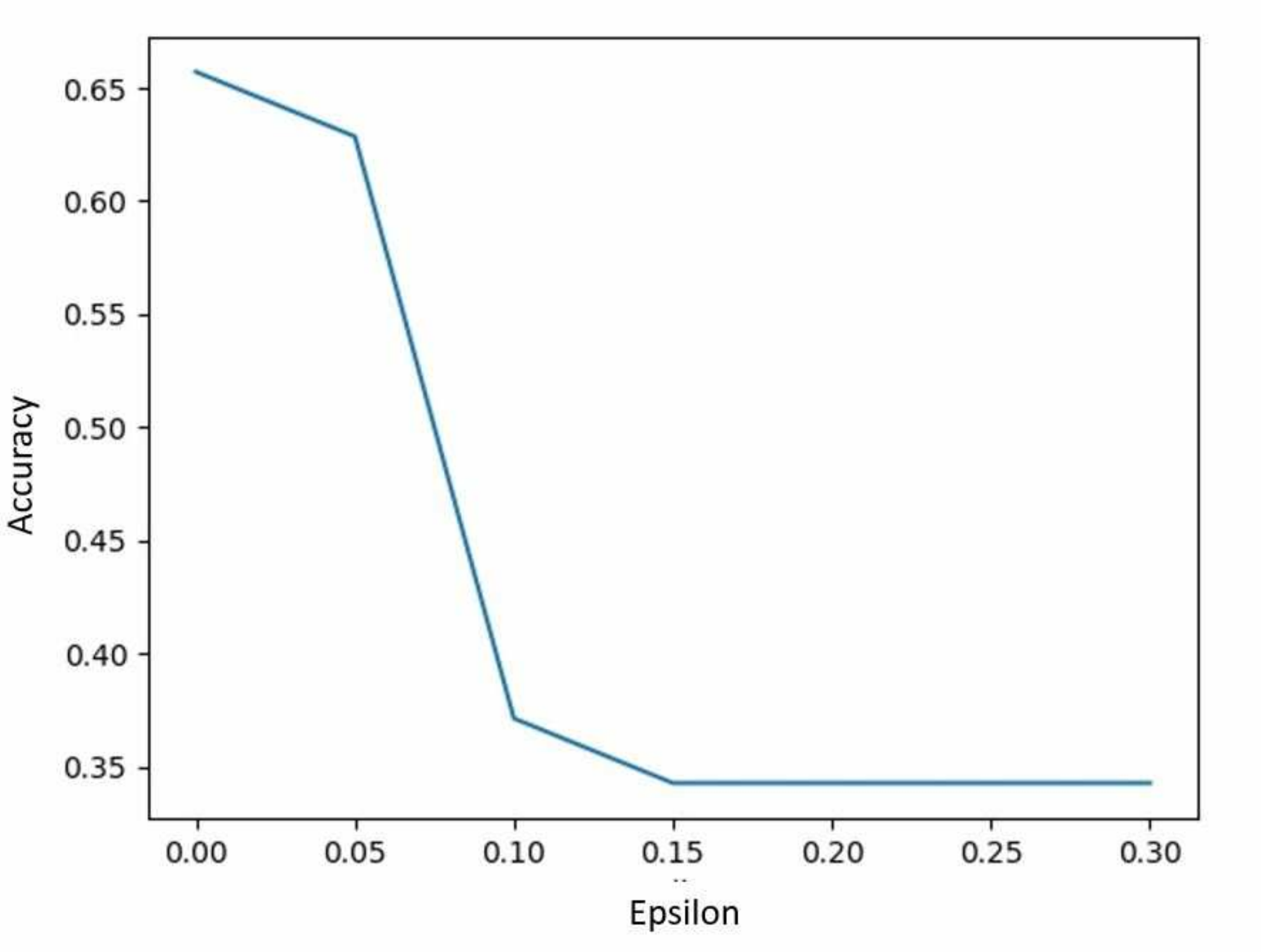}
    \caption{Accuracy vs. Penetration Coefficient with High Epsilon Values.}
    \label{accuracy_high}
  \end{minipage}\hspace{.3cm}
  \begin{minipage}[b]{.48\textwidth}
    \includegraphics[width=\textwidth]{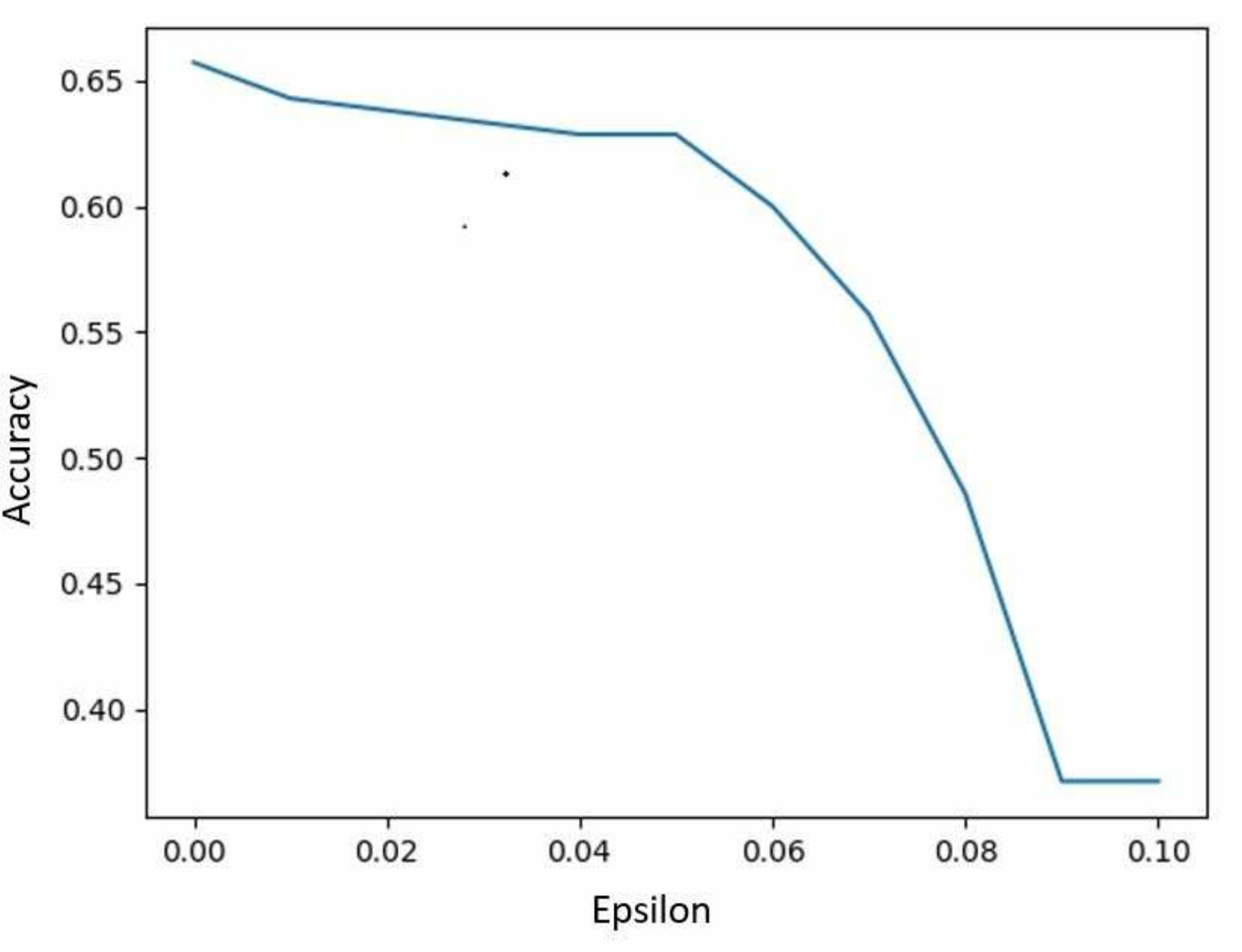}
    \caption{Accuracy vs. Penetration Coefficient with Small Epsilon Values.}
    \label{accuracy_small}
  \end{minipage}
\end{figure*}

When we train our model, the model's accuracy reach to 70 with clean image. The reason of the higher performance value is not found is because being trained with a small dataset. It is believed that this number would be higher if the model was trained with a bigger dataset. In this study, we emphasize how a CNN classifier is vulnerable against adversarial attack rather than enhance the model's performance. Graphics show how the model's performance is compromised with different penetration coeficients in Figure \ref{accuracy_high} and in Figure \ref{accuracy_small}.

\begin{figure*}[htbp]
  \begin{minipage}[b]{0.48\textwidth}
    \includegraphics[width=\textwidth]{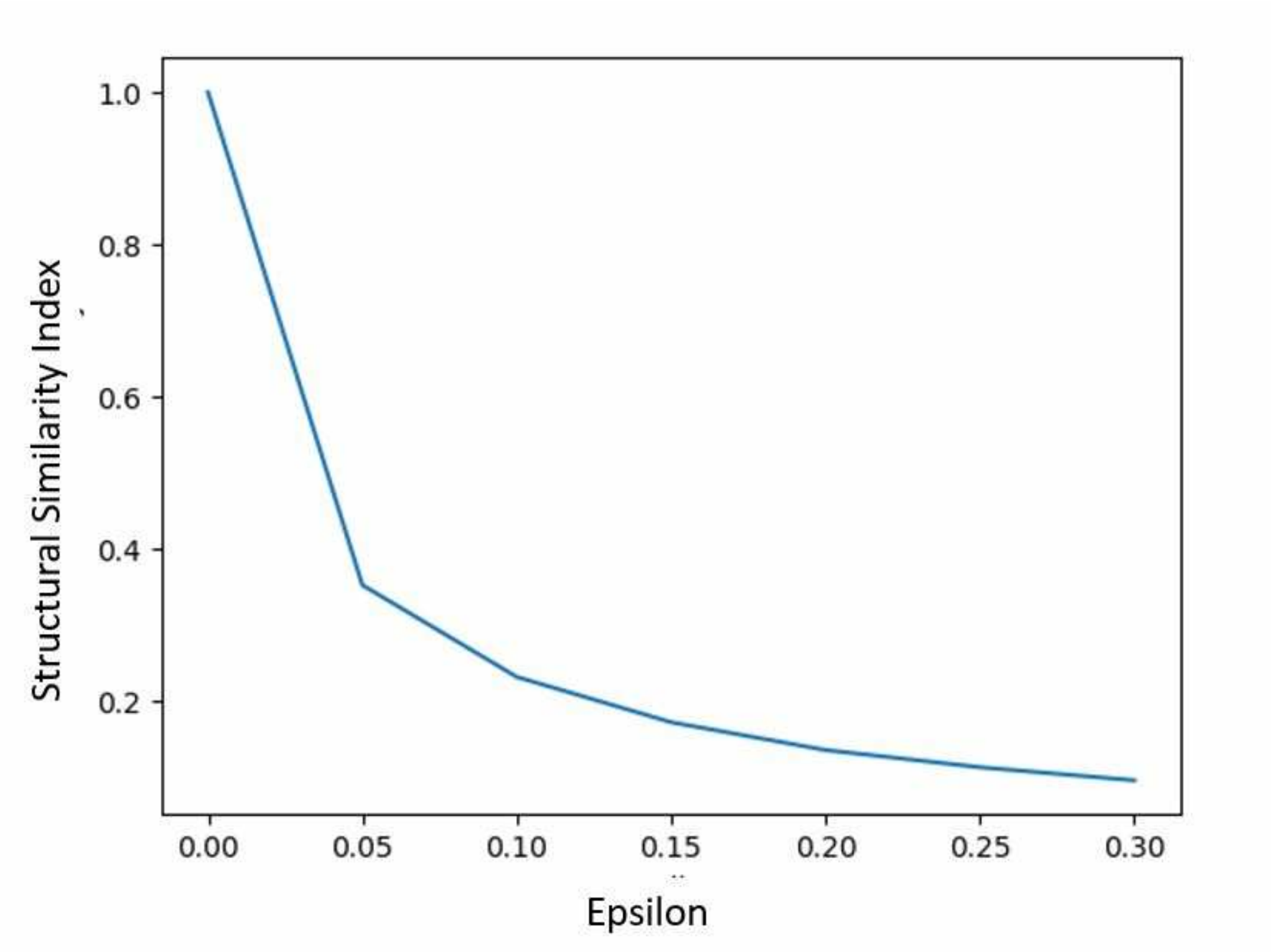}
    \caption{Structural Similarity Index vs. Penetration Coefficient with High Epsilon Values.}
    \label{similarity_high}
  \end{minipage}\hspace{.3cm}
  \begin{minipage}[b]{.48\textwidth}
    \includegraphics[width=\textwidth]{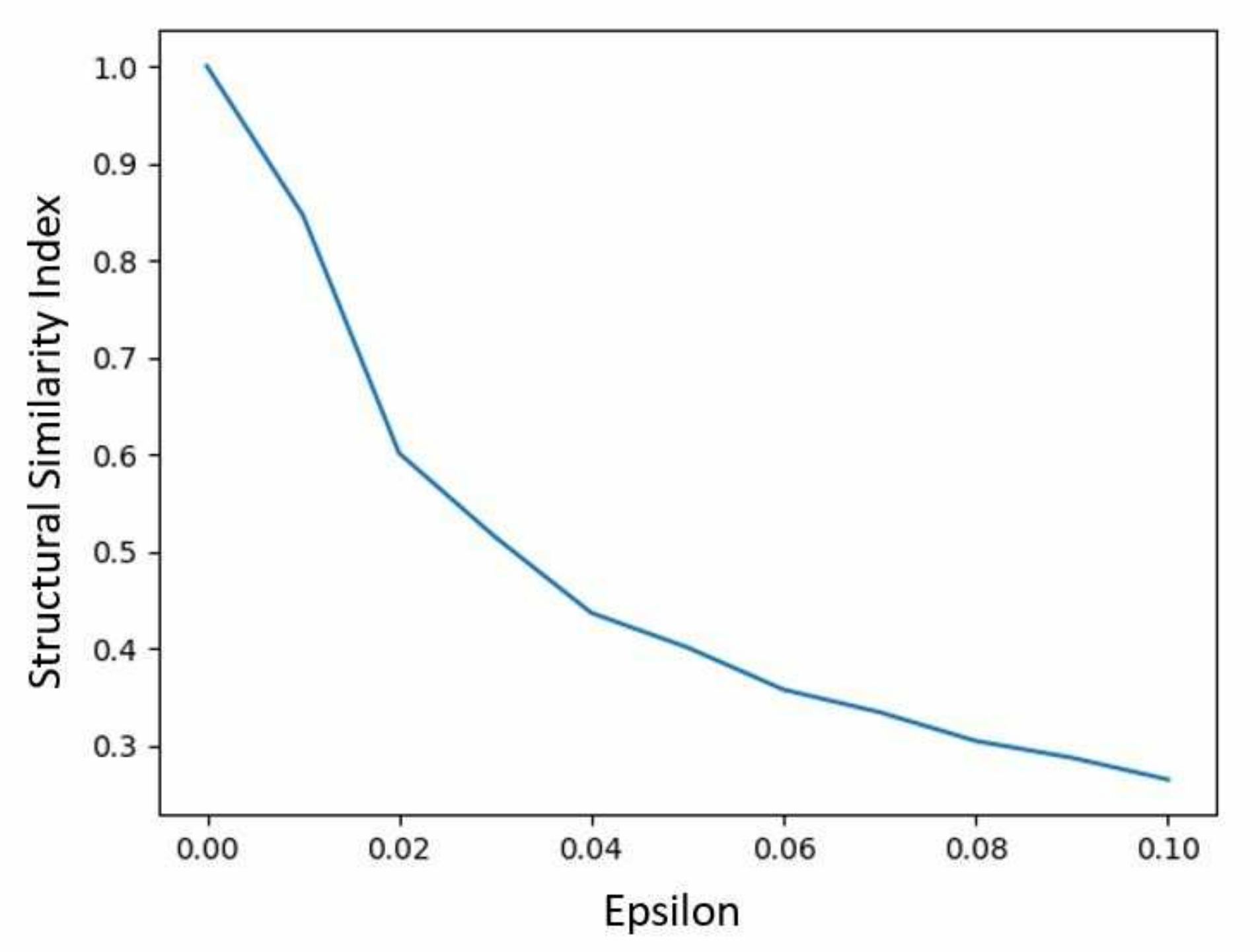}
    \caption{Structural Similarity Index vs. Penetration Coefficient with Small Epsilon Values.}
    \label{similarity_small}
  \end{minipage}
\end{figure*}

Also, we observe how penetrated images are degenerated by different epsilon values using SSIM. Graphics illustrate these differences in Figure \ref{similarity_high} and in Figure \ref{similarity_small}.

\section{Related Work}
Adversarial attacks against DNN classifiers were discussed previously by different researchers. in this research, we carry out a FGSM attack against a mammographic image classifier based on CNN to compromise its performance first time. Szegedy et al. \cite {szegedy2015going} showed various of intriguing attributes of neural networks and relevant models. Malicious samples which they produced were very close to original examples that the human eye perceives as identical. They performed on ImageNet dataset and their adversarial examples were misclassified by several classifiers with different architectures. Goodfellow et al. \cite {goodfellow2014explaining} introduced FGSM attack for the first time. They applied their attack on MNIST dataset \cite {deng2012mnist} including handwritten digits from 0 to 9 and achieved to mislead classifiers. They attempted to maximize error by adding optimum noise. 

Papernot et al. \cite {papernot2016limitations} address an issue as well and applied Jacobian-based Saliency Map Attack (JSMA). They attempted to find most important pixel features which lead to significant changes to the output by modifying in a way that is imperceptible. Moosavi-Dezfooli et al. \cite {moosavi2016deepfool} also discussed adversarial attack . They identified the decision boundary of deep learning classifier based on inputs and tried to find closest distance to the decision boundary pixels of the original input which were manipulated to misguide the model. They implemented adversarial attacks against only deep learning classifiers. Therefore, they called it a deepfool attack. Then, they improved their deepfool adversarial attack and they formulated to find universal penetration vector in \cite {moosavi2017analysis}.

Nguyen et al. \cite {nguyen2015deep} designed a different adversarial attack using evolutionary algorithms (EAs) in which they managed to fabricate images to misclassify a high confidence classifier in an optimum way. Carlini and Wagner \cite {carlini2017adversarial} created a novel adversarial attack called a cws attack as the initial letter of their names . They claimed that their attack type could evade all existing adversarial detection defenses at this time when they published their works. 

Su et al. \cite {su2019one} aproached adversarial attacks from a different point of view and managed to produce adversarial samples by altering one pixel value of original image. Rozsa et al. \cite {rozsa2016adversarial} generated great numbers of adversarial attack samples from one original image by transformation or rotation of that image. Zhao et al. \cite {zhao2017generating} generated adversarial samples synthetically using generative adversarial network (GAN) instead of transformation or rotation of an original image. In GAN concept, there are two different classifiers named discriminator and generator. Generator first creates a random noise and tries to capture data distribution of the original image. Discriminator takes original images and images fabricated by generator as input and distinguishes between them and give feedback to generator. Generator always tries to fabricate better images which can not be distinguishable by discriminator. At one point generator's synthetic images can not be recognized by discriminator. Thus, new samples can be created. \footnote{Other works have studied the problem
\cite{baza2019b,parksmarnet,baza2019blockchain,CC3,BC2,baza2019detecting,Baza3,firmware2,omar2,baza2019sharing,parkccnc,omar1,shafee2019mimic}}

\section{Conclusion}
NN is used a lot for breast cancer detection and classification with high accuracy. In this research, we discuss the security and vulnerability of CNNs. We present an attack algorithm to exploit the security gap of CNNs in breast cancer classification and raise awareness of radiologists and doctors. The adversarial strategy is viable without changing the architecture of the algorithm, however, in order to solely modify input data. We evaluate our adversarial attack with different penetration coefficients and show these changes in a mathematical way using structural similarity index. Hopefully, our work will promote further research to increase the robustness of CNN in breast cancer classification. Although there have been much research regarding defense strategy against adversarial attacks, countermeasures for detection of malicious samples have not been well-studied and are still a significant problem. Currently, we are working to find adapted threshold value to differentiate real image from fake one. In this way, intrusion samples can be detected automatically.

\bibliographystyle{IEEEtran}
\bibliography{CC} 
\let\mybibitem\bibitem
    
\end{document}